\documentclass[twocolumn]{el-author}

\newcommand{\ua}{\uparrow}
\newcommand{\nc}{\newcommand}
\nc{\da}{\downarrow} \nc{\hc}{\hat{c}} \nc{\hS}{\hat{S}}
\nc{\bra}{\langle} \nc{\ket}{\rangle} \nc{\eq}{equation (\ref}
\nc{\h}{\hat} \nc{\hT}{\h{T}}\nc{\be}{\begin{eqnarray}}
\nc{\ee}{\end{eqnarray}}\nc{\rd}{\textrm{d}}\nc{\e}{eqnarray}\nc{\hR}{\hat{R}}\nc{\Tr}{\mathrm{Tr}}
\nc{\tS}{\tilde{S}}\nc{\tr}{\mathrm{tr}}\nc{\8}{\infty}\nc{\lgs}{\bra\ua,\phi|}\nc{\rgs}{|\ua,\phi\ket}
\nc{\hU}{\hat{U}}\nc{\lfs}{\bra\phi|}\nc{\rfs}{|\phi\ket}\nc{\hZ}{\hat{Z}}\nc{\hd}{\hat{d}}\nc{\mD}{\mathcal{D}}
\nc{\bd}{\bar{d}}\nc{\bc}{\bar{c}}\nc{\mc}{\mathcal}\nc{\ea}{eqnarray}\nc{\mG}{\mathcal{G}}\nc{\bce}{\begin{center}}
\nc{\ece}{\end{center}}
\date{12th December 2011}

\begin{document}

\title{An Efficient Algorithm for the Piecewise-Smooth Model with Approximately Explicit Solutions}

\author{Huihui Song, Yuhui Zheng, Kaihua Zhang}

\abstract{This paper presents an efficient approach to image segmentation that approximates the piecewise-smooth (PS) functional in~\cite{12} with explicit solutions. By rendering some rational constraints on the initial conditions and the final solutions of the PS functional, we propose two novel formulations which can be approximated to be the explicit solutions of the evolution partial differential equations (PDEs) of the PS model, in which only one PDE needs to be solved efficiently. Furthermore, an energy term that regularizes the level set function to be a signed distance function is incorporated into our evolution formulation, and the time-consuming re-initialization is avoided. Experiments on synthetic and real images show that our method is more efficient than both the PS model and the local binary fitting (LBF) model~\cite{4}, while having similar segmentation accuracy as the LBF model.}

\maketitle

\section{Introduction}

Active contours or snakes have been attracted much attention in image segmentation, and numerous methods from edge-based models~\cite{1,3,5,7,11} to region-based ones~\cite{2,4,9,10,12} have been proposed. Therein, the Chan and Vese (CV) model~\cite{2} is one of the most popular region-based models, which aims to look for a particular partition of the image that represents the object and the background, respectively. It can be taken as a particular case of minimal partition problem of the Mumford and Shah model~\cite{6} for image segmentation, and the level set method proposed by Osher and Sethian~\cite{7} has been successfully explored to implement it.
In~\cite{12}, by adopting multiple level set functions to represent multiple regions, Vese and Chan further extended their CV model to multiphase segmentation. Since the CV model assumes the intensities inside and outside the contour to be statistically homogeneous, it cannot work well for images with intensity inhomogeneity. To address this issue, in~\cite{12}, Vese and Chan also proposed the other model that utilizes the piecewise smooth functions to fit the image intensity, which achieves better performance for the images with intensity inhomogeneity. Tsai et al.~\cite{10} also proposed a similar method, contemporaneously and independently. These models are called piecewise smooth (PS) model. The PS model utilizes smooth functions to accurately approximate image intensities, thereby being able to segment the images with intensity inhomogeneity satisfactorily. However, they need to iteratively solve three coupled PDEs simultaneously with very expensive computational cost. Besides, to reduce the computational cost, the initial contour should be set close to the object boundaries. For example, the method in~\cite{10} uses the CV model to achieve a preliminary segmentation. However, for the image with intensity inhomogeneity, the initial contour obtained from the above method is still far away from the desired object boundaries, and the computational cost is still expensive. Furthermore, the extensions of the piecewise smooth functions of the PS model are difficult to be performed in practical application. In addition, re-initialization is necessary in implementing the PS model with the level set method, which also increases computational burden.

To address these issues, Li et al.~\cite{4} proposed an effective model termed as the local binary fitting (LBF) model, which performs favorably well on segmenting images with intensity inhomogeneity. LBF achieves satisfying segmentation results as the PS model, but with much more efficient performance. However, at least two convolutions with a large kernel must be computed in each iteration in LBF, thereby increasing its computational cost.

In this paper, motivated by the Scale-space theory~\cite{8}, by setting some rational constraints on the initial conditions and the final solutions of the PS model, we propose two novel formulations which can be approximated as their explicit solutions. Only one PDE needs to be solved in our method, thereby reducing the computational cost significantly. Moreover, a penalized energy functional for regularizing the level set function is incorporated into the PS energy functional and re-initialization can be eliminated. Complexity analysis reveals that our method is much more efficient than LBF~\cite{4}.

\section{Principle of Our method}
\label{sec:sec2}
From the theory of Scale-space~\cite{8}, it is well-known that the evolution of a function according to its Laplacian is equivalent to filtering the initial version of the function with a Gaussian function whose standard deviation is related with the evolution time:
\begin{equation}
I(\mathbf{x};t)=I_0(\mathbf{x})\otimes K_t(\mathbf{x}),
\label{eq:dffusion}
\end{equation}
where $I_0(\mathbf{x})$ denotes the original image intensity, and $K_t(\mathbf{x})$ is the Gaussian kernel with the standard deviation $t$. $I(\mathbf{x};t)$ is the solution of the heat conduction or diffusion equation as follows
\begin{equation}
\frac{\partial I}{\partial t}=\Delta I,
\end{equation}
with the initial condition $I(\mathbf{x};t=0)=I_0(\mathbf{x})$, where $\Delta$ denotes the Laplacian operator.

In the PS model~\cite{12}, a smooth function $u^+(\mathbf{x})$ is explored to fit the image intensity inside the contour $C=\{\mathbf{x}:\phi(\mathbf{x})>0\}$, which is obtained by solving the following equation
\begin{equation}
\frac{\partial u^+}{\partial t}=\Delta u^+-\frac{1}{\mu}(u^+-IH(\phi))~\mathrm{in} \{\mathbf{x}:\phi(\mathbf{x})>0\}.
\label{eq:pspluseq}
\end{equation}
As we use the smooth function $u^+$ to approximate $IH(\phi)$, it is rational to set the initial condition of (\ref{eq:pspluseq}) to be $u^+(\mathbf{x};t=0)=aIH(\phi)$, where $a$ is a constant, which results in the final solution $u^+\approx IH(\phi)$. Obviously, the solution we obtain can be approximated as that of the following equation
\begin{equation}
\frac{\partial u^+}{\partial t}=\Delta u^+~\mathrm{in} \{\mathbf{x}:\phi(\mathbf{x};t)>0\}
\label{eq:oureq}
\end{equation}
with the initial condition $u^+(\mathbf{x};t=0)=aIH(\phi)$. Note that the extension formulation of $u^+$ in the PS model~\cite{12} is the same as (\ref{eq:oureq}). Therefore, we only need to solve the following equation in the whole image domain $\Omega$
\begin{equation}
\frac{\partial u^+}{\partial t}=\Delta u^+~\mathrm{with}~u^+(\mathbf{x};t=0) = aIH(\phi),
\label{eq:explicit}
\end{equation}
whose explicit solution is as
\begin{equation}
u^+(\mathbf{x};t) = K_t\otimes(aIH(\phi))=aK_t\otimes(IH(\phi)).
\label{eq:solution}
\end{equation}
Considering the final solution of (\ref{eq:explicit}) can be approximated as $u^+\approx IH(\phi)$, we can obtain
$a\approx\frac{IH(\phi)}{K_t\otimes(IH(\phi))}$. Moreover, it is rational to assume that the image intensity in the local region is homogeneous~\cite{4}, so the term $K_t\otimes(IH(\phi))\approx(K_t\otimes H(\phi))IH(\phi)$ because we utilize a truncated Gaussian kernel to approximate $K_t$ as in~\cite{4}. Therefore, the constant $a$ in (\ref{eq:solution}) can be approximated as
\begin{equation}
a\approx\frac{1}{K_t\otimes H(\phi)}.
\end{equation}
Thus, the solution of (\ref{eq:oureq}) in the whole image domain $\Omega$ can be approximated as
\begin{equation}
u^+\approx\frac{K_t\otimes(IH(\phi))}{K_t\otimes H(\phi)}.
\label{eq:approxsolution}
\end{equation}
Similarly, we can obtain the fitting function $u^-$ outside the contour $C$ in the whole image domain as
\begin{equation}
u^-\approx\frac{K_t\otimes(I(1-H(\phi)))}{K_t\otimes (1-H(\phi))}.
\label{eq:approxsolutionmius}
\end{equation}

By putting (\ref{eq:approxsolution}) and (\ref{eq:approxsolutionmius}) into the evolution function of the PS model, we only need to solve the level set evolution formulation, so our method is much more efficient than the PS model. Furthermore, we incorporate a signed distance regularized term~\cite{5} into the level set evolution formulation to reduce the expensive re-initialization procedure, and the total formulation of our method is as follows
\begin{equation}
\begin{split}
\frac{\partial\phi}{\partial t}&=\alpha\left(\Delta\phi-\nabla\left(\frac{\nabla\phi}{|\nabla\phi}\right)\right)+\delta(\phi)[v\nabla\left(\frac{\nabla\phi}{|\nabla\phi|}\right)-|u^+-I|^2-\mu|\nabla u^+|^2\\
&+|u^--I|^2+\mu|\nabla u^-|^2],
\end{split}
\label{eq:ourls}
\end{equation}
where $\alpha>0, v>0$ and $\mu>0$ are fixed constants. $\delta(\cdot)$ is the Dirac function.
\section{Advantages over PS and LBF}
Different from the PS model~\cite{12}, our method does not need to solve the coupled PDEs to obtain the fitting functions $u^+$ and $u^-$, and it is unnecessary for our method to extend $u^+$ and $u^-$ to the whole image domain since they have been defined on the whole domain. Moreover, the level set function in our method can be simply initialized to be constants with different signs inside and outside the contour, and re-initialization is unnecessary due to the distance regularization term in our method.

The main computational cost of the LBF model is to compute the term $\lambda_1e_1-\lambda_2e_2$ in its evolution functions, which can be written as follows for efficiency
\begin{equation}
\begin{split}
\lambda_1e_1-\lambda_2e_2&=(\lambda_1-\lambda_2)I^2(\mathbf{x})(K_{\sigma}(\mathbf{x})\otimes \mathbf{1})\\
&-2I(\mathbf{x})(K_{\sigma}(\mathbf{x})\otimes(\lambda_1u^+-\lambda_2u^-))\\
&+K_{\sigma}(\mathbf{x})\otimes(\lambda_1u^{+2}-\lambda_2u^{-2}).
\end{split}
\label{eq:e1e2}
\end{equation}
Different from our method that only needs to compute the term $K_{\sigma}(\mathbf{x})\otimes \mathbf{1}$ once when computing $u^-$, the LBF model needs to compute the other two convolutions in (\ref{eq:e1e2}) in each iteration, which makes it less efficient than our method.
\section{Results}
\label{sec:sec3}
All the partial derivatives $\frac{\partial\phi}{\partial x}$ and $\frac{\partial\phi}{\partial y}$  in (\ref{eq:ourls}) are approximated by using the simple finite difference scheme in~\cite{5}. We truncate $K_t$ to be an $m\times m$ mask for efficiency, where $m$  is the smallest odd number more than $4t$, in which $t\in[1,2]$. We apply our method to synthetic and real images of different modalities with the same parameters $\alpha=0.02$,$v=0.001\times 255^2$, $\mu=0.02$, and the time-step $\Delta t=0.025$.
\begin{figure}[t]
\begin{center}
\includegraphics[width=0.9\linewidth]{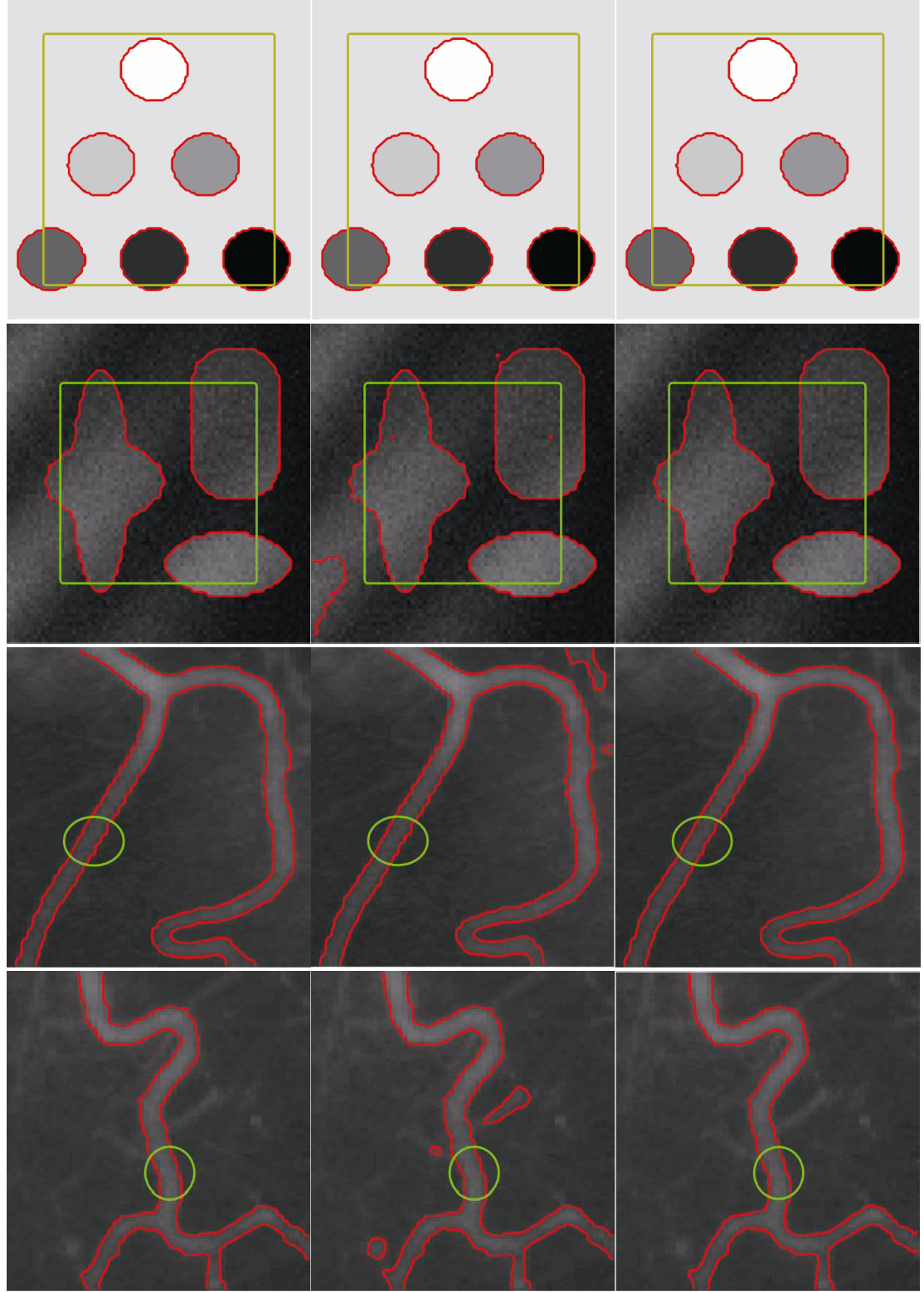}
\end{center}
   \caption{Comparisons of our method, the PS model~\cite{12}, and the LBF model~\cite{4}. The left column: results of our method. The middle column: results of the PS model. The right column: results of the LBF model. The initial contours are highlighted in green color while the final segmentation contours are highlighted in red.}
\label{fig:results}
\end{figure}

Fig.~\ref{fig:results} shows the results of our method, the PS model, and the LBF model. As we can see from the left column and the right column, respectively, the results of our method and the LBF model are almost the same, which validates that our method can sufficiently utilize the local image information, and obtain the accurate segmentation results. The middle column shows the results of the PS model, as for the objects with intensity inhomogeneity (see from the second row to the fourth row), the main objects are extracted satisfactorily, but some unwanted contours also appear, and the final contours are somewhat noisy.
\begin{table}
\begin{center}
\begin{tabular}{|l|c|c|c|c|}
\hline
Name & Image1 &Image2 &Image3 &Image4 \\
\hline\hline
PS~\cite{12} & 2000.15&850.13&1911.12&2100.12\\
LBF~\cite{4}&30.42&16.36&18.01&18.38\\
Ours & 5.23 &1.22&3.77&4.02\\
\hline
\end{tabular}
\end{center}
\caption{CPU cost (in seconds) for PS, LBF, and our method.}
\label{table}
\end{table}
Table.~\ref{table} shows the CPU cost of our method, PS, and LBF for the images in Fig.~\ref{fig:results} in the same order. The validation is implemented in Matlab 2010a on a 2.8-GHz Intel Pentium IV, 4G RAM personal computer on Windows system. Obviously, our method is much more efficient than the PS model. Furthermore, although the LBF model is much efficient than the PS model, it is less efficient than our method.
\section{Conclusion}
\label{sec:sec4}
In this paper, we have proposed two novel formulations which can be approximated as the explicit solutions of the coupled PDEs in the PS model, so only one PDE needs to be solved, and its computational cost has been reduced significantly. Comparisons on synthetic and real images with the PS model and LBF model show the advantages of our method in terms of efficiency over the PS and LBF models, while having the similar visual accuracy as the LBF model. It should be pointed out that the proposed formulations can be easily extended to the Fourth-phase PS model in~\cite{12}.
\vskip3pt
\ack{Huihui Song is supported in part by the Natural Science Foundation of China under Grant
41501377, in part by B-DAT (Nanjing University of Information
Science and Technology) under Grant KXK1406, and in part by the Foundation of Jiangsu Province of China under Grants  15KJB170012 and BK20150906.. Yuhui Zheng  is supported by the Natural Science Foundation of China under Grant 61402235. Kaihua Zhang is supported by the Natural Science Foundation of China under Grant 61402233, and in part by the Foundation of Jiangsu Province of China under Grant BK20151529.}

\vskip5pt

\noindent Huihui Song(\textit{Jiangsu Key Laboratory of
Big Data Analysis Technology (B-DAT) and Jiangsu Collaborative Innovation Center on Atmospheric
Environment and Equipment Technology (CICAEET), Nanjing University of Information
Science and Technology, Nanjing, China})\\
 Yuhui Zheng (\textit{School of Computer and Software,
Nanjing University of Information Science and Technology, Nanjing, China})\\
\noindent Kaihua Zhang(\textit{Jiangsu Key Laboratory of
Big Data Analysis Technology (B-DAT), Nanjing University of Information
Science and Technology, Nanjing, China})\\
\noindent E-mail: zhkhua@gmail.com\\

\end{document}